\documentclass{article}
\usepackage{ijcai16}

\usepackage{times}

\title{Defining Concepts of Emotion: From Philosophy to Science}
     \author{Changqing Liu    \\
             ClinTFL Ltd., UK \\
             C.Liu@ClinTFL.com
            }

\begin{document}

\maketitle

\begin{abstract}

This paper is motivated by a series of (related) questions as to whether a computer can have pleasure and pain, what pleasure (and intensity of pleasure) is, and, ultimately, what concepts of emotion are.

To determine what an emotion is, is a matter of conceptualization; namely, understanding and explicitly encoding the concept of emotion as people use it in everyday life. This is a notoriously difficult problem (Frijda, 1986; Fehr \& Russell, 1984). This paper firstly shows why this is a difficult problem by aligning it with the conceptualization of a few other so called semantic primitives such as ``EXIST", ``FORCE", ``BIG" (plus ``LIMIT"). The definitions of these thought-to-be-indefinable concepts, given in this paper, show what formal definitions of concepts look like and how concepts are constructed. As a by-product, owing to the explicit account of the meaning of ``exist", the famous dispute between Einstein and Bohr is naturally resolved from linguistic point of view. Secondly, defending Frijda's view that emotion is action tendency (or Ryle's behavioral disposition (propensity)), we give a list of emotions defined in terms of action tendency. In particular, the definitions of pleasure and the feeling of beauty are presented.

Further, we give a formal definition of ``action tendency", from which the concept of ``intensity" of emotions (including pleasure) is naturally derived in a formal fashion. The meanings of ``wish", ``wait",  ``good", ``hot" are analyzed.

\end{abstract}

\section{\textbf {Introduction}}

To explain what emotions are, philosophers and psychologists put forward various theories. Many observations of theirs capture important aspects of emotion phenomena; e.g. many agree that desire is a component of emotions. In particular, Ryle observes that emotions are dispositions to behave. Frijda defines emotion concept as ``action tendency". Wierzbicka uses ``want" as a basic term to define emotion concept. The building blocks of their theories includes such concepts as ``tendency", ``satisfaction", ``reward", ``punishment"( Frijda, 1986), disposition (Ryle, 1949), ``unexpectness" (Ortony, Clore and Collins, 1988), ``want", ``good", ``bad" (Wierzbicka, 1992) and ``motivation", ``desire", ``goal" or ``purpose", etc. These building concepts, called semantic primitives, are thought to be indefinable. Thus the emotion theories stop at the level of these indefinable concepts. So without (formal) definition of what people mean by ``action tendency", ``want", ``good", ``satisfaction" etc, none of the theories can convince people such and such thing in a computer is the equivalent of ``action tendency", ``want" etc.; any robot claimed to have emotion would always be thought to be emulating the manifested behavior of emotions, rather than emotions themselves.

The reason for the lack of the definitions of these semantic primitives may be two fold. First, traditionally (before AI) there have been rarely need to formally define the meanings which are intuitively clear. People including philosophers are happy with daily language in expression, comprehension and arguments about everyday, ``commonsense" world around them; there is no problem at all in people's usage of them in daily life (this fact reminds us of Wittgenstein's view that the meaning of a word is its use). Even the meaning of ``force" in physics has not been defined. Only two exceptional occasions are: a) when people were asked to find the meaning of ``limit" after it had already been employed by mathematicians for 100 years, and b) when Einstein felt need to investigate what ``simultaneously" really mean. Second, the task to find the ``real meaning" is difficult, as in the cases of ``simultaneously", ``limit", ``big", ``exist". As Nilsson (1991) pointed out

\textit {`` ... there are some particularly difficult subjects to conceptualize. Among these are … processes, events, actions, beliefs, time, goals, intentions, and plans  ...  Interestingly,  many of the most difficult conceptualization problems arise when attempting to express knowledge about the everyday, ``commonsense" world. AI researchers join company with philosophers who have also been attempting to formalize some of these ideas. Choosing to use first-order predicate calculus as a representation language does not relieve us of the chore of deciding what to say in that language. Deciding what to say is harder than designing the language in which to say it!"} \footnote{Indeed, the definition of ``limit" was first described in natural language by Cauchy. No doubt it is a formal definition.}

In this paper, we show by examples how the semantic primitives are defined in a formal manner. Having analysed the concept of force and tendency, we define ``action tendency" formally and thus ground the concepts of emotion on more fundamental formal concepts. With the definition of emotion, we are allowed to give the answer in coherent manner to the question of what emotion intensity is. \footnote{This problem has rarely been addressed in previous literature}.

Our approach to meaning of semantic primitives including ``tendency" has the same flavour as Cauchy's to the meaning of ``limit" which is intuitively clear while to find the definition took people more than 150 years. The approach may be summarized as ``concepts are constructed out of the observation of behavioral regularities".

\section{\textbf Defining concepts (formally)}
We will give several examples to show what concepts' (formal) definitions look like and to demonstrate why it is not easy to define a concept formally.

The first example is \textbf{``BIG"}. This concept is intuitively clear. To define it is not a trivial problem \footnote{'BIG' is in Goddard's (2002) list of semantic primitives}.  Obviously, 'big' actually means 'bigger (than)' because by 'big' people mean something bigger relative to other things. Without loss of generality, we consider 'longer' for simplicity. So what is the meaning of ``stick A is longer than stick B"? The meaning of ``longer" is "whenever you align one end of A with one end of B and do cut along the other end of A, you will see an extra piece of stick off B"; ``cut" here represents any operation of comparing. In other words, the interpretation of the string ``$x > y$'' is $x - y > 0$ which in turn is interpreted as   ``whenever you make a comparison, you will see one piece of A, one piece of B (of the same size), and another piece of stick off B".  Concept is constructed out of observations of behavioral regularity.

%I planned to take the concept ``circle" as another example. After scrutiny of how this concept, among others (e.g. "straight line"), is formed, we admitted that it should be viewed as a different class of concepts. The concepts of this class are constructed through "pattern recognition", whereas those this paper focuses are constructed out of observation of behavioral regularity -- ``behavioral regularity recognition".

\textbf{``FORCE"}. Believe or not, the concept of ``force" has no definition so far, though people understand what it means intuitively.  The following situation is one of many from which people form their concept of ``force". You carrying a tray with one of your hands, whenever (every time, at any time) you withdraw your support, the tray will fall down to the ground. Let us write the concept of force as ``$ \neg  C  \to  M$" where C stands for ``condition''. The interpretation of it is, whenever there is no support (i.e. the condition $C$ does not holds) movement $M$ occurs. This is the definition of ``force'', we suggest. In the definition, the behavioral regularity is signified by the notion ``whenever ...". Further, you may have a another feeling of another force such that whenever you don't use both hands to support the tray it falls down, namely, ``$ \neg  (C_1 \land C_2)  \to  M$" ($C_1$ denotes that there is support of the first hand and $C_2$ the second hand). Thus you acquire the concept of ``intensity" of force; you distinguish between force A and force B by their condition parts, i.e.  $\neg C_1$ vs $\neg (C_1 \land C_2)$, so that you say ``B is stronger than A" in this sense. In physics, only properties of force such as magnitude and direction, rather than the concept itself, are dealt with.\footnote{Perhaps it is not necessary to examine the everyday concept of force for the purpose of physics. But it is not always unnecessary to examine the everyday concepts for the purpose; for example, the concept of ``simultaneity".} Neither has the semantic interpretation of force ever been defined in linguistics. We are not concerned with real force in the real world, unlike Hume. We are just interested in the concept of ``force", a behavioral regularity.

\textbf{``LIMIT" (``infinitesimal", ``infinity")}. It is not difficult for school students to understand what $\lim \frac{1}{n} \to 0$ means. In folk language, it means $\frac{1}{n}$ approaches to $0$ as $n$ goes bigger and bigger indefinitely. It is not difficult for school students to understand what this means. They apprehend it well as Newton, Leibniz and Euler did. However, to translate it into a formal statement is so difficult that scholars struggled for 100 years before finding its formal definition, $(\varepsilon, \delta)$-procedure (thanks to Cauchy and Weierstrass). It can be seen that the $(\varepsilon, \delta)$-procedure definition does not resemble its intuitive counterpart at all. This fact might illustrate why defining the concept of limit formally is not an easy task. In fact, when high school students enter universities and are exposed to the formal definition of limit all of them are confused by the formal definition; it is strange and different from limit learning in high school. Not only does it take long time for them to understand the formal definition of the limit, but also almost all of them grasp the formal definition only in the sense that they can remember it and correctly apply it to solve problems given. Indeed, it is difficult for them to understand why $(\varepsilon, \delta)$-procedure is the meaning of ``limit''. This fact demonstrates where the difficulty of conceptualization stems from; the formal definition looks so different that it is hard for people to link them together. The same can be said with respect to the case of emotion. \footnote {The second difficulty of emotion is body-mind problem, which however is solved by Gilbert Ryle to mention one.}

\textbf{``EXIST/THERE IS"}. People have an intuitively clear idea of what we are saying by ``there is/exists a table in the room". There is no definition for its meaning \footnote{This is also in Goddard's (2002) list of semantic primitives }. Formally, we suggest, when you say ``there is/exist a table'' the meaning of this alphabetical string is no more than, whenever you ``look'' at there you will ``see'' it. This is the definition of ``exist''. A human being gets this object permanence concept when he is as young as an infant (Bremner, 1994) and, we argue, it won't change substantially when he becomes a toddler, a teenager, an adult and a scientist like Einstein.

%[Although there might be object permanence, we people can even never be able to raise such a question as to if it really exists, because any question involving such word `really exist' or the like is meaningless, unless you mean ``whenever ... you ...''. ]

Like in the example of `big', here `look' and `see' just represent any way of perception (look, touch, smell ...) and its outcome (true or not) respectively. People often say, ``because the moon exists, whenever we look we will see it''. This saying misleads people into thinking that `existence' and the behavior regularity ``whenever ... we will see it'' are cause and effect; two separate things. The truth is the reverse, namely that it is because you observe the ``whenever ... you see'', you construct the concept of ``existence''. Philosophers debate whether there is a real existence which is independent of people's observation. We don't want to involve this type of philosophical argument in this paper but just wonder curiously that  what the meaning is in their exposition. We agree with those maintaining that that metaphysical thing is beyond the capability of scientific concept. Scientific theory must be meaningful; every concept must be meaningful - associated with a given interpretation.

Let's turn to Bohr--Einstein debate. Einstein's famous challenging question is ``Is the moon there when no one looks''? To both Bohr and Einstein, if the answer is ``no'', then this contradicts the commonsense. If the answer is ``yes'', then it means there is an absolute reality (existence) which is independent of people's observation, and this is inconsistent with quantum theory's observation. Unable to answer Einstein's question directly, quantum theorists eschewed the type of debates and they ``shut up and calculate''. To be sure Bohr is proved right and Einstein is wrong. However, Einstein's question is still there. Here we suggest a direct answer to Einstein's question. Einstein's question can be translated as ``If no one looks, is it true that whenever one looks one can see a moon?''. So the hypothesis that no one looks does not refute the proposition ``whenever one looks one sees it''. In other words, Einstein's question can be expressed in notion ``if $\neg A, A \to B$ = True?" \footnote{unless his 'exist' is of metaphysics, and therefore, we maintain, is meaningless}. Obviously, the answer is ``Yes''. It is Einstein who should ``shut up". Or he must give the meaning of his `exist' explicitly; had he tried to do so, we believe, he should have found its real meaning which he grasped when he was a toddler, what we make explicit here.

%DavidYerle.com discusses 'existence'.

\section {\textbf {Emotion is desire}}
Gilbert Ryle (1949) see emotions as being behavioral dispositions. Frijda (1986) defines emotion as action tendency. Wierzbicka (1992) characterizes emotions in terms of wanting to do, to mention a few. It can be seen that this group of philosophers tend to eliminate mental terms from accounts of emotion. Their theories are convincing, at least for some of us AI researchers. In this paper, we will give formal definitions for action tendencies and thereby show how computers can have emotions. To make the reduction from the commonsense idea about emotion to the formalized concept mildly, we reduce the concepts of emotion to desires firstly. If the readers already accept Frijda's action tendency account for emotion, they can skip the first half of this section and turn to the part of pleasure and feeling of beauty. Indeed, intuitively it seems that desiring to do conveys the same meaning as tendency to do, or, action tendency (or, a state of readiness to execute a given kind of action) ---can you interpret explicitly the difference in meaning, if you feel it at all?  They are virtually synonymous. With regard to the idea that emotions are equivalent to desire, much previous work has provided insight on this.  Hebb (1949, p. 190) defines hunger as the tendency to eat.  Amold (1960) says that emotion is felt action tendency.  Searle (1984, p. 24) says that to be thirsty is to have, among other things, the desire to drink. The difference between their theories and ours lies in that we identify emotion with desire conceptually --- namely every emotion is simply a desire (to do ...). Feeling hungry is identified with the desire to eat, an itch with the desire to scratch, feeling thirsty with the desire to drink, and so on. It is often said ``because I am hungry, I desire to eat". Again this form of phrasing misleads people into thinking that "being hungry" and "desire to eat" are two separate things. When we are talking of thirst we are, by that very token, talking of the desire to drink.  Or, it can be said, the desire to drink causes thirst, in the sense in which we speak of one thing explaining, or being explaining of, another thing. The second reason of resorting to the notion ``desire" is for the sake of syntax; it would be more natural to say "desire to do" than "action tendency to do". The following is a mini dictionary of emotions defined by desires.

\setlength{\hangindent}{30pt}

       \textit{feeling hungry: desiring to eat;}

       \textit{feeling thirst: desiring to drink;}

       \textit{feeling an itch: desiring to scratch; for instance, feeling an itchy toe is the desire to scratch the toe.}

       \textit{feeling cold (hot): desiring to warm(cool) oneself;}

       \textit{feeling of fear: desiring to flee, or escape, etc.; that I feel fear of someone/something means I desire to go away from it.}

       \textit{feeling of love: desiring to be with;}

       \textit{regret: desiring to do what could have made oneself not do what was done by oneself;}

       \textit{hate: desiring to make someone feel pain, die, etc.; desiring to retaliate someone;}

       \textit{pain(in the general sense of unpleasant feeling): desiring, when P (a fact or a state of affairs) is true,
       to do something as a result of which P will not hold; }

       \textit{feeling of excitement (social emotion): desire to cheer;}

       \textit {pleasure: desiring to do what is being done. (More about this later.); }

\noindent
For instance, desiring to warm myself when I am warming myself (making heat flow into the body)--- for when I am warming myself by, say, sitting near a fireplace, my body  may physically still need heat flowing into it---is  a kind of pleasure.  Similarly, (when one is drinking) desiring to drink (we do not call it thirst in this case) is a kind of pleasure. In this fashion, every particular emotion should be defined in terms of desire (to do ...). A common mistake is to suppose that pleasure results after one's thirst is quenched. For another example, desiring to scratch oneself when one is doing so is an instance of pleasure.  (It is undeniable that when one is scratching an itch one is feeling a kind of pleasure.)  Where there is no desire there is no pleasure. It can be said that pleasure is awareness of ``desire to do A when you are doing A".

To determine what an aesthetic feeling is, is considered a difficult problem.  The aesthetic feeling of beauty, as it is commonly considered, refers to the pleasant feeling caused by seeing a beautiful object such as a flower, a girl, etc..  But what is ``beautiful" then? In fact, the sense in which an object is ``beautiful" is that whenever one sees the object in question one has a feeling of beauty.  Therefore, the notion ``beautiful" ought to, as many have pointed out, be eliminated.  Actually, the adjective ``beautiful" makes no reference to any objective property of the object, but simply that:  when seeing the object one has a pleasant feeling.  We have the following definition:

      \textit { feeling of beauty : desiring to see the object while I am actually seeing it. }

\noindent
Thus, feeling of beauty is an instance of pleasure, the pattern of ``one is doing what one desires to do" in concept. Feeling of the sublime is another puzzle to philosophers. In short, when you contemplate a picture of scenery, which in a real life situation would induce a  feeling of fear, it is to be said that you have a feeling of the sublime, which is a sort of pleasure, as aestheticians put it. With the above defined concept of pleasure, here we explain what sublime is.

Imagine this painted scene; under the black cloud-covered sky, a broad expanse of the sea darkened to gray, no ship sailing.  If you were on the sea all alone, you would feel fear.  Bear in mind, this is a painting, a work of art.   We argue that while you are looking at the picture without telling yourself (by shifting your consciousness to the fact that you are now in a room of the gallery.) that you are not really on a sea but just in a gallery, you are, at that very point, seeing the sea as if you were really on it: Or, to put it another way, you have the same experience of seeing the sea as if you were really on it.  As a result, you have a feeling of fear---desire to flee, or, to get away from the scene. Immediately afterward, you have escaped: You are now aware of being safe. Then you look at the picture again. Contemplating the picture is a procedure which consists of the repeated almost instantaneous occurrences of
begin:
      seeing the sea
      desiring to get away
     having got away
end.

As it can be seen, at a macro-level, there is a form of desiring to do A when doing A, namely desiring to get away when getting away, in this process, which is recognized as pleasure of a sort, an instance of pleasure defined above. In real life, when you are really on the sea in question, you would in no way have a sense of having gotten away from it, and simply a feeling of fear. This might be, and, in fact, is just what Burke says: ``when danger or pain press too nearly, they are incapable of giving any delight, and are simply terrible;"

To conclude, the study of emotions should be the study of desires rather than, as some psychologists have tended to do, to explore physical and physiological causes of emotions.  The approach to this problem, we suggest, is to examine what one desires to do when having an emotion.  To determine what emotions are is to determine what desires are.  The only reason an emotion is claimed not a desire is that either the desire has not been correctly identified or that feeling is not an emotion.

\section {\textbf {Formal definition of action tendency (disposition, propensity)}}

As mentioned above, at the semantic level we treat desire and action tendency equally. Therefore, sometimes we use 'desire' for 'action tendency' for the sake of convenience. Frijda (1986) construed ``action tendencies'' as ``states of readiness to execute a given kind of action, [which] is defined by its end result aimed at or achieved'' (p 70)``.  His interpretation uses such notions as ``readiness'', ``achieve'' and ``aim'',  the meanings of which are not clear. For us, ``action-tendency'' sounds clearer than the interpretation given by him. We now begin our analysis of the concept of ``action-tendency'' (desire).   Consider the following cases:

Case 1: a horse tied to a stake, making an effort to move forward, the reins being drawn tight, Everything is still.

Case 2: a bow fully drawn with the arrow loaded; everything is still.

Case 3: a stone put on a bridge; everything is still.

\noindent
   Then, compare the following:

Case a: the horse tied to the stake, standing calmly; everything is still.

Case b: the bow not drawn, with the arrow loaded; everything is still.

Case c: the stone put on the bank, beside the bridge; everything is still.

\noindent
It cannot be denied that there is a kind of thing which case (1), case (2) and case (3) all have, while case (a), case(b) and case(c) do not.  What is it, or, how should we express it?  Allow us to say, it is a kind of tendency.  But what is a tendency?  One might say that that thing is just a physical force.  Admittedly, it is force.

This leads to the definition of action tendency: $S \to A$, where $S$ is a circumstance or fact, which can be described by a propositional formula, and $A$ is action or occurrence of event.\footnote{It would be more precise to say that $S$ names a type of circumstance, and $A$ a type of event. For the sake of convenience, without loss of points, we will let the present usage stand.} The interpretation of $S \to A$ is whenever $S$ holds $A$ happens, the same form as the concept of ``force".  So we say that tendency means just $S \to A$. It cannot be said that tendency causes $S \to A$, but should be said that we signify $S \to A$ by the word ``tendency".  Both force and desire are tendencies. Returning to the cases described above, in case (1), case (2) and case (3), $S$ is respectively: the reins are cut, the bow string is released, the bridge is broken.  And $A$ is respectively: the horse running forward, the arrow shooting, the stone dropping down.  In case (1), for example, when the reins are cut, the horse moves, while in case (a), it won't.

In the case of feeling hungry, $S$ may be this: food is at hand, or food is in my mouth, or food is at hand and at the same time I am not on a diet, and so on, and the a is eating.  If the case of thirst, $S$ may be this: water, or beer, and the like is at hand, etc.  The action $A$ characterizes the desire's quality, which determines a desire's characteristic feel or quale.  For example, if $A$ is eating, the desire $S \to A$ counts as feeling hungry, and if scratching, counts as feeling an itch, and so forth.  The condition $S$ determines a desire's intensity, of which we will go into details in the next section.

Since action tendency is now defined formally, definition of pleasure given above is a formal definition, without any mental term, a scientific concept. If one programmed a robot to have a mechanism of $S \to A$, one would be right to claim that one had produced an emotion, because one had produced what is fundamentally identical to what people designate as ``emotion".

%; one more instance of pleasure: desire to urinate when one is urinating.

%It might be questioned that how we could know that an agent---if it really has a desire but the condition $S$ has never held, and besides will in no way hold so that no action is performed--- has the desire. For example, if it is true that I am itching but in practice take no action to scratch, how the desire could be said to exist.  This question amounts to asking  how to realize of an agent an internal state which is described as $S \to A$.

\section {Intensity of action tendency (emotion intensity)}

%Frijda thought the action tendency could not account for emotion intensity. In his book %(1986), only one page focuses on the notion ``emotion intensity'', denying the concept
%though.
Our daily language and experience recognize that emotions may exist in varying degrees of intensity.  For example, ``I am feeling a little bit hungry", ``I am feeling hungry" and ``I am feeling very hungry", do reflect the conception of emotional intensity; in essence the three emotions are simply three mentally distinguishable feelings, which, though distinguishable, cause the same action ---namely, eating.  We often distinguish between apprehension, anxiety, fear, and panic or between contentment, pleasure, happiness, joy and ecstasy.  Furthermore, there seems to be a kind of order relationship among emotions.  People are clearly aware that if emotion $E_1$ is "stronger" than $E_2$, and $E_2$ is "stronger" than $E_3$, then there is the same relationship between $E_1$ and $E_3$ as between $E_1$ and $E_2$, or between $E_2$ and $E_3$ ---transitivity.  And people would not think that emotion $E_2$ is "stronger" than $E_1$. If $E_1$ is "stronger" than $E_2$ , that, for example, a bit hungry is "stronger" than very hungry---antisymmetry.  Also, people do not think that there is such a relationship between an emotion and itself as that between, say, very hungry and a bit bit hungry---antireflexivity.  Yet  strangely enough psychologists and philosophers have rarely examined this dimension of variation of emotion (Plutchik, 1991,p. 14). Like most psychologists, Frijda denies emotion intensity as a unitary concept (p. 290). Instead, they treat emotion intensity as the stimulus intensity, or variation in some dimensions of emotional behavior. In this stance, \textit {being very hungry} vs \textit {being a little bit hungry} may be measured by having not eaten for ten hours vs five hours. This sounds not to describe the hunger itself, but one of the causes of it. People do not mean ``very hungry" in that sense.

%In fact, as it will be seen, the action-tendency definition for emotion is so coherent, more than Frijda himself thought, that it captures the intensity dimension of emotion.

Our theory of emotion allows us to coherently address this problem. The formal definition of emotion naturally captures the dimension of emotional intensity .

The key to the problem of emotion intensity, we view, is to elucidate the concept of emotional intensity, to determine what the meaning of ``stronger" is.
To begin with, it is instructive to reflect on some important facts which might be easily overlooked:

%10      In practice, a proposition naturally determines that a type of circumstance, rather than an individual  circumstance, must be the case in order to make it true; for example, if P= a cup of liquid is at hand and it is   drinkable then all such circumstances as a cup of beer is at hand, a cup of tea is at hand, etc., would make P hold true.  For the sake of simplicity, here we would not distinguish between the notions of ``proposition"   and ``propositional formula".

%11    The fact that P holds true means that the agent believes that p.  For reasons of space, we shall not address the issue of  ``belief" here.  Instead, we just give a definition: The agent believes that p iff this very P holds true.

\textbullet\ When extremely hungry, one would eat whatever edibles are available; while if only a bit hungry, one would not until some ``delicious" food is available.

\textbullet\ When feeling a mild need to urinate, one might urinate only when a convenient opportunity presents itself; while if feeling an urgent need to urinate, one might urinate without so much (or any) attention to the place and its convenience.

\textbullet\ With feeling a little bit hungry and at the same time very thirsty, if both food and drink are available one would begin to drink rather than eat.

\textbullet\ et cetera.

\noindent Further probing what it means for an emotion to be stronger than another leads to the following definition.

Let $E_1$ be $S_1 \to A_1$ and $E_2$ be $S_2 \to A_2$.  Using $E_1 > E_2$  for ``$E_1$ is stronger than $E_2$", we have

\textbf{Definition (stronger)}:  $E_1 > E_2$  iff  If $E_2$ gives rise to action $A_2$ in some circumstances then $E_1$ will necessarily give rise to action $A_1$ under the same circumstances, and not vice versa; in other words, there cannot be such a circumstance that $A_2$ is triggered under it while $A_1$ is not, but there are some circumstances under which $A_1$ will be triggered whereas $A_2$ is not.

For example, suppose $A$ is eating, then $ C_1  \to A $ is stronger than $ C_1 \land C_2 \to A $; namely,  $hunger_1$ is stronger than $hunger_2$. $C_1 \lor C_2 \to A $ is stronger than $C_1 \to A $. The condition part of action tendency determines a desire's intensity.

Therefore, to have a strong pleasure simply means satisfying (executing $A$ of the desire) a strong desire (thus 'satisfaction' is clearly defined). %Then the question is how to generate stronger desire in an agent when, e.g., feed him a piece of music?

Since the intensity of emotion or desire only involves the condition parts of emotions or desires, two emotions of different quality may be comparable with each other with respect to intensity.  This is also reflected by our intuitions.  For example, it seems meaningful to say that feeling a little bit hungry is weaker than feeling extremely thirsty.  Nevertheless, not every pair of emotions are comparable with one another.  For example, it seems that, intuitively, some feelings of fear and some of thirst are, to some extent, not comparable with one another, with respect to intensity.  This is because, suppose $E_1$ and $E_2$ are not comparable with one another, neither $E_1 > E_2$   nor $E_2 > E_1$  holds, in this case. The conditions specified by the condition parts not only refer to the external world states but also may include the internal states of an agent.  For example, when one feels hungry, one might take no action to eat even if some food is at hand, if one is on a diet; with the hunger becoming strong enough, one would eat the food.

\section {More about Pleasure and Pain}

As has been seen, the concept of emotion is stood for by $S \to A$, and thereby we mean that an emotion, or desire, is uniquely determined by a condition and an action. Pleasure has not this characteristic form of expression.  In fact, if we call those that are characterized as the form $S \to A$, such as fear, itch, etc., ``basic" or ``fundamental" emotions, pleasures are not basic emotions; they are ``forms" of desires, rather than desires themselves.  As mentioned earlier, the concept of pleasure is defined as ``desire to do $A$ while $A$ is being performed". For instance, one desires to scratch (that is, one is itchy) when one is scratching.

This is best illustrated by examples.  Suppose the desire now is thirst, namely desire to drink water.  Let us write $S(\cdot)(w)$ to represent the condition part of the thirst, and $A$ to represent the action---drinking.  The notation of $S(\cdot)(w)$  is used to indicate that the condition under which $A$ occurs depends on the level of fluid content of one's body or in one's stomach, or throat.  The lower the level is the weaker the condition is and therefore the stronger the desire to drink is, and vice versa.  In short, the condition part $S(\cdot)$  is a function of the level of fluid content.
The occurrence of $A$ can have affects on $S(\cdot)(w)$, leading to the increase of the level of the fluid content and this in turn causing $S(\cdot)(w)$  to vary.  After one's having taken a swallow of water, the thirst would be weakened.  The goal of $A$, and also of the desire, is, say, ``water is flowing through the throat".  At the starting point, the desire is

$S(\cdot)(w0) \to A$.

\noindent
When it comes that $S(\cdot)(w1)$ is still satisfied by a circumstance---for example, the one that water is at hand---$A$ occurs, that is, it occurs taking a swallow of water.  Thereby, the level of fluid content would be up to a higher point, $w1$, and thus the thirst now is

$S(\cdot)(w1) \to A$.

\noindent
Providing $S(\cdot)(w1)$ is still satisfied by the circumstance, the second swallow of water would be taken.  And so forth.  This procedure terminated at a point when, because enough water has flown into the body, $S(\cdot)(w)$ is too strong to be satisfied so that, we say, the intensity of thirst is ``zero", that is, one now does not feel thirsty.  During the whole process, one keeps having the desire to drink, and meanwhile keeps drinking; provided $S(\cdot)(w)$ is true, nothing would cause one to stop drinking.  As it is, the more intense the thirst is, the more intense the pleasures experienced when drinking are.  Neither occurrence of action nor desire are themselves pleasures; as has been emphasized, in order to characterize a pleasure there must be mention of both a desire and the occurrence of the action which is desired. Phenomenologically, there are no such distinct emotions as ``pleasures".

  One might argue that pleasures result from satisfaction---that desires are satisfied.  Disregarding the issues of the notion of ``satisfaction", if ``desires are satisfied" refers to the fact that the goal of a desire holds---that is, thus interpreted, the goal is a state of affairs (for example, the fluid content of one's body is at a normal level)---then why the stat that the fluid content of one's body is on a normal level (therefore one does not feel thirsty, i.e., does not desire to drink) does not cause one to feel pleasure, even though this state of affair was the goal of one's desire an hour ago when one was thirsty?

  Traditionally it is believed that desires and pleasures are persistent, i.e., they exist in an interval rather than at a point of time.  But essentially, ``a fact holds during an interval" means no more than that the fact in question holds every time we test it.  Consider the proposition ``$ S $ holds during the interval [t1 ,t2]".  What is the meaning of the proposition? It can but mean that every time (during the interval) you test $ S $  you, at that point, obtain a truth.  When it is not tested, it is neither a truth nor falsehood, and it is meaningless to say that it is at that point true or false. Therefore when a notion is presented that something holds during an interval, it coherently refers to the case where something holds repeatedly as opposed to ``continuously" during an interval. This allows us to say that the repeated occurrence of desire to flee and action of fleeing (see the early account of aesthetic feeling), counts as pleasure --- the characteristic of aesthetic feelings. Parenthetically, to account for the functionality of pleasure, we argue that pleasures' function is to keep one doing (or repeatedly doing) what one is doing (or has done).

When one is having a pain, one has a desire to do an action which will cause a certain fact not to hold; without taking into account the difference between the two forms of goals---one is a proposition, another is in the form of the negative of a proposition---we can simply say that pain is a desire to do an action which has not taken place.  This explains why feeling thirsty is a painful, love is a sort of pain, and so on.

\section {\textbf{``WISH'' and ``WAITING"}}  ``Desire to do'' vs ``wish $G$ happens'' signify the different meanings of the two notions. The latter is usually seen in the scenario when a person can do nothing to achieve $G$ while waiting for it to happen. For example, as an audience, a person wishes Manchester United to score a goal. Our analysis of the concept of 'wishing' and 'waiting' gives the definition of wish as follows. When you are waiting a present sent by a friend of yours, the wish is `` $G$ (the present arrives) $\to$ stop checking''. In the case of watching football game, the wish is ``$G$ $\to$ stop checking and start to cheer (or desire to cheer)". The meaning of `waiting' can be signified by the difference between such two circumstances: a) that one is sitting there without any wish---he is not waiting, and b) that sitting there with a wish --- he is waiting. That is to say, at the semantic level and in this context (one may wait bad news not wished), `waiting' and `wishing' are equivalents. That is why wish cannot be defined in terms of waiting, without circular. Wishing has not got the same definition of intensity as desire. However, people say ``I have a stronger wish". What does that mean? That means more `frequently checking' or ``a strong desire to cheer when the goal is achieved". Therefore, in the case of wishing, the frequency or strength of the action part of the definition is defined as the strength of a wish. Nevertheless, the condition part of the definition can still be recognized as one dimension of intensity, as desire. For example, ``$ G \to$ Stop waiting'' is stronger than ``$S \lor G \to$ Stop waiting'', because in the situation of the former the agent keeps checking until only $G$ is true, whereas in the latter $S$ (say ``it is raining") can stop the checking.

\section {Meaning of ``good"}

When one says a thing is good, one is merely expressing that one likes it, rather than referring to something objective of the thing. When one voices ``something is good", the object (usually, the speaker) to whom it is good is tacitly referred to. Again, it is not because it is good that I want it, but vice versa. There is no something ``good" but people's desire to have it. This is another example that inappropriate linguistic forms are expressed in statements whose superficial grammatical form mistakenly engenders the hypostasization of non-existent objects of various sorts. The delicious food is delicious just because we have a strong desire to eat it, rather than because of the ``deliciousness". The water is $80\,^{\circ}\mathrm{C}$ refers an objective of water, whereas to say the water is hot refers to your own desire. The same argument applies to 'beauty'.

\section {\textbf{Conclusion}}

To determine what an emotion is, is to find a scientifically defined concept of emotion, which accords with our
everyday concept of what an emotion is. Hence what we should do is to make clear the everyday concept -- to
express it explicitly. Conceptualization is difficult to make when attempting to express knowledge about the everyday, ``commonsense" world.  Consequently, the existing theories stop at the level of ``semantic primitives" which are thought indefinable. In this paper, we have shown how to divide those `atoms' into smaller ``particles"; we demonstrate how semantic primitives can be further defined in a formal manner.

\section*{Acknowledgments}
The author would like to thank Spencer Nash for helpful discussion and his suggestion about the strategy of the paper writing of such a challenging topic.

%% The file named.bst is a bibliography style file for BibTeX 0.99c
%\bibliographystyle{named}
%\bibliography{ijcai16}

\end{document}